\PassOptionsToPackage{svgnames}{xcolor}
\documentclass[letterpaper, sigconf, 10pt]{acmart}
\usepackage[T1]{fontenc}
\usepackage[utf8]{inputenc}
\usepackage{amsmath, cite, url}
\usepackage{graphicx}
\usepackage{hyperref}
\usepackage{natbib}
\usepackage{adjustbox}
\usepackage{rotating}
\usepackage[]{tikz}
\usetikzlibrary{calc}

\setcopyright{none}
\renewcommand\footnotetextcopyrightpermission[1]{}
\title{\huge{Simulation of Neural Responses to Classical Music\\Using Organoid Intelligence Methods}}

\author{Daniel Szelogowski}
\email{dszelogowski@gmail.com}
\affiliation{
  \institution{Capitol Technology University}
  \streetaddress{Street Address}
  \city{Laurel}
  \state{MD}
  \country{}
  \postcode{Post Code}
}

\begin{document}

\begin{abstract}
Music is a complex auditory stimulus capable of eliciting significant changes in brain activity, influencing cognitive processes such as memory, attention, and emotional regulation. However, the underlying mechanisms of music-induced cognitive processes remain largely unknown. Organoid intelligence and deep learning models show promise for simulating and analyzing these neural responses to classical music, an area significantly unexplored in computational neuroscience. Hence, we present the PyOrganoid library, an innovative tool that facilitates the simulation of organoid learning models, integrating sophisticated machine learning techniques with biologically inspired organoid simulations. Our study features the development of the Pianoid model, a ``deep organoid learning'' model that utilizes a Bidirectional LSTM network to predict EEG responses based on audio features from classical music recordings. This model demonstrates the feasibility of using computational methods to replicate complex neural processes, providing valuable insights into music perception and cognition. Likewise, our findings emphasize the utility of synthetic models in neuroscience research and highlight the PyOrganoid library's potential as a versatile tool for advancing studies in neuroscience and artificial intelligence.
\end{abstract}

\keywords{Artificial Intelligence (AI), Organoid Intelligence (OI), EEG Response, Organoid Learning, Computational Neuroscience, Neuromorphic Computing, Music Perception and Cognition}

\maketitle
\pagestyle{plain}

\section{Introduction}
\label{sec:introduction}
The emerging field of \textbf{Organoid Intelligence (OI)} represents a novel intersection of stem cell technology, bioengineering, and artificial intelligence, offering a platform for exploring complex neural phenomena in vitro \citep{Smirnova_Caffo_Gracias_Huang_Moralesetal._2023}. This emerging field utilizes three-dimensional brain organoids (derived from human pluripotent stem cells) to replicate aspects of human brain function, including neurogenesis, synaptic activity, and network formation. The integration of these organoids with advanced computational methods \textemdash{} such as deep learning networks \textemdash{} enables the simulation and analysis of neural processes, thus providing a unique opportunity to study organoid learning and adaptive responses to various stimuli, including auditory inputs.

In \textbf{Music Perception and Cognition (MCP)}, a field that studies the cognitive psychology and mechanisms relating to music \citep{Justus_Bharucha_2002}, investigating neural responses to classical music has provided significant insights into the brain's ability to process complex auditory information \citep{Koelsch_2011}. Classical music, with its rich harmonic structures and dynamic range, serves as an ideal stimulus for studying cognitive functions such as attention, emotional regulation, and memory. The analysis of \textbf{electroencephalogram (EEG)} data during music listening has revealed distinct patterns of brain activity, shedding light on the neural mechanisms underlying auditory perception and the emotional impact of music \citep{Daly_Williams_Hwang_Kirke_Miranda_Nasuto_2019}.

Deep learning approaches have revolutionized the analysis of EEG signals, allowing for more nuanced and detailed interpretations of neural data \citep{Roy_Banville_Albuquerque_Gramfort_Falk_Faubert_2019}. These methods enable the extraction of complex features from EEG recordings, facilitating the study of how different brain regions interact during music perception. By applying deep learning techniques to EEG data obtained from subjects listening to classical music, researchers can uncover patterns associated with specific cognitive and emotional responses, enhancing our understanding of the neural basis of music perception and cognition. This paper thus aims to leverage the capabilities of OI and deep learning to simulate and analyze these neural responses, providing new insights into the cognitive processes elicited by classical music, as well as presenting a new Python library for the simulation of (what we term as) \textbf{Organoid Learning (OL)} models.

\vspace{-0.2cm}
\subsection{Motivation, Goals, and Objectives}
The motivation behind replicating EEG responses to classical music listening in an organoid model stems from the desire to bridge the gap between human cognitive neuroscience and artificial intelligence. Traditional EEG studies on music perception have provided invaluable data on the brain's response to complex auditory stimuli. Still, these studies are limited by the invasive and complex nature of in vivo experiments. By developing organoid models that can simulate these neural responses, we aim to create a more accessible and ethical platform for studying the underlying mechanisms of music perception and cognition. This approach allows for more controlled experimentation and allows the potential to explore neurodevelopmental and neurodegenerative conditions in a more precise and replicable manner.

Furthermore, the introduction of a new Python library for simulating Organoid Learning models addresses a critical need in the field of \textbf{biocomputing}. This library facilitates the creation and customization of specialized organoids, tailored environments, and simulation schedulers, enabling researchers to implement \textbf{Machine Learning (ML)} and \textbf{Deep Learning (DL)} models that interface with organoid systems. The ability to simulate learning processes within organoids provides a powerful tool for testing and refining AI algorithms in a biologically relevant context. This innovation both enhances the fidelity of computational models in reflecting human-like neural processes and accelerates the development of biocomputing applications, such as \textbf{personalized medicine} and adaptive learning systems. Hence, the overarching goal is to harness the potential of OI to advance our understanding of complex neural phenomena and to create new pathways for integrating biological and artificial intelligence.

\section{Related Work}
\label{sec:related_work}
This section reviews key research on EEG responses to music, the integration of AI and organoid intelligence, and the implications of biocomputing for understanding neural dynamics. These studies provide a foundation for exploring the potential of OI in simulating complex neural processes in response to auditory stimuli.

\subsection{Ogata, 1995}
Ogata \citep{Ogata_1995} investigated the psychophysiological effects of classical music on human EEG, comparing it with a simulated white noise condition that mirrored the sound-pressure variations of the music. Eight volunteers had their EEG, ECG, and EOG recorded under three sound conditions: silence, classical music, and simulated music (sim-music). The study found that during drowsiness (stage S1), the power densities of the delta EEG component were higher with sim-music than with classical music, suggesting higher levels of physiological consciousness when listening to music. Participants reported feeling relaxed and comfortable with music but were weary and sleepy with sim-music, indicating a significant difference in mental set and consciousness between the two sound conditions. The study highlighted the complexity of music's components \textemdash{} rhythm, melody, and harmony \textemdash{} and how their absence in sim-music led to different physiological and psychological responses, reflecting music's unique ability to maintain higher levels of consciousness and organized mental activity even in transition states from wakefulness to sleep. The experiment used Tchaikovsky's ``Swan Lake'' as the classical music and its sound-pressure-matched sim-music, observing EEG responses in stages W1 (awake/alert), W2 (awake/relaxed), S1 (NREM stage 1), and S2 (NREM stage 2). The results showed that sudden increases in loudness in music did not always lead to increased consciousness, unlike sim-music, where changes in loudness did \textemdash{} implying that the musical context affects how changes in loudness are perceived and processed by the brain. Subjective reports of comfort and alertness with music and discomfort and sleepiness with sim-music accompanied the physiological EEG differences between the music and sim-music conditions. These findings suggest that music's characteristic effects on consciousness and mental activity are linked to its complex acoustic structure, which maintains attention and engagement more effectively than white noise. This study provides evidence for the potential of music as a tool for maintaining higher levels of consciousness, suggesting further investigation of its therapeutic applications for sleep and mental disorders.

\vspace{-0.2cm}
\subsection{Daly et al., 2019} 
Daly et al. \citep{Daly_Williams_Hwang_Kirke_Miranda_Nasuto_2019} conducted a study examining the neural responses to music-induced emotions by employing a joint \textbf{EEG} and \textbf{functional magnetic resonance imaging (fMRI)} approach. The researchers aimed to understand the relationship between EEG-based neural correlates of approach-withdrawal behavior and activity changes in the subcortical emotional response network. The study involved participants listening to music designed to elicit a range of affective states while recording their EEG and fMRI data. Participants continuously reported their felt affective states, allowing researchers to analyze the covariation of self-reports with EEG and subcortical brain activity. The results indicated that prefrontal EEG asymmetry reflects cortical activity and changes in sub-cortical regions such as the amygdala, posterior temporal cortex, and cerebellum. Additionally, the study found that the magnitude of the EEG asymmetry was correlated with the activity of the limbic and paralimbic systems. In contrast, the entropy of the asymmetry was associated with the autonomic response network, particularly the auditory cortex.

The study highlights the potential of using EEG to monitor changes in sub-cortical brain activity induced by music, suggesting its applicability in music therapy. The findings revealed that EEG prefrontal asymmetry is a reliable indicator of sub-cortical changes in activity, which are typically considered ``invisible'' to EEG. Daly et al. established a significant link between music-induced emotional states and neural responses by demonstrating that both generated and classical music listening can modulate activity in critical regions of the limbic system. This research underscores the complexity of music as a stimulus, given its dynamic temporal structure and ability to induce rapid changes in affective responses. Furthermore, it supports the use of music as a unique tool to investigate affective responses and their neural foundations, particularly in therapeutic contexts to treat emotional disorders. Thus, this research provides a comprehensive view of how the cortical and subcortical regions interact during music-induced emotional experiences, bridging the gap between the EEG and fMRI modalities.

\subsection{Cai et al., 2023} 
Cai et al. \citep{Cai_Ao_Tian_Wu_Liu_Tchieu_Gu_Mackie_Guo_2023} propose a novel AI hardware approach using brain organoids, termed Brainoware, to address the limitations of traditional silicon-based AI systems. Traditional AI hardware, based on silicon chips, faces challenges such as high energy consumption, heat generation, and limited capacity due to the \textbf{von Neumann bottleneck}. In contrast, the human brain, with its efficient \textbf{Biological Nural Networks (BNNs)}, serves as an inspiration for developing more efficient AI hardware. Brain organoids, derived from human pluripotent stem cells, can self-organize into 3D structures mimicking brain tissue, offering advanced complexity, connectivity, and neuroplasticity. Brainoware utilizes these properties by interfacing brain organoids with a high-density multielectrode array to perform computations. The study demonstrates that Brainoware can process spatiotemporal information, achieve nonlinear dynamics, and exhibit fading memory, enabling unsupervised learning by reshaping the organoid’s functional connectivity. This is illustrated through applications such as speech recognition and predicting nonlinear equations, where Brainoware adapts to input signals via synaptic plasticity. Despite promising results, challenges remain in organoid generation, maintenance, and data management. Improvements in \textbf{organoid interfacing} and optimization of data handling are necessary for advancing this technology. This research suggests that integrating brain organoids into AI hardware could significantly enhance computational efficiency and learning capabilities, providing a new paradigm for AI development.

\subsection{Smirnova et al., 2023}
Smirnova et al. \citep{Smirnova_Caffo_Gracias_Huang_Moralesetal._2023} introduce the emerging field of \textbf{Organoid Intelligence (OI)}, which leverages human stem cell-derived brain organoids to replicate critical molecular and cellular aspects of learning, memory, and cognition in vitro. The concept of OI aims to establish a multidisciplinary field that utilizes brain organoids as \textbf{biological computing systems}, harnessing their capabilities through advanced bioengineering and scientific methods. These brain organoids, characterized by high cell density, enriched glial cells, and gene expression critical for learning, are cultivated with integrated microfluidic perfusion systems and novel 3D microelectrode arrays to support spatiotemporal chemical signaling and high-resolution electrophysiological recording. This setup enables the exploration of brain organoids' potential to recapitulate the molecular mechanisms of learning and memory, facilitating the development of biocomputing models that rely on stimulus-response training and \textbf{organoid-computer interfaces.} The envisioned applications include creating networked interfaces where brain organoids connect with real-world sensors and output devices and other organoids to perform complex tasks using biofeedback, big-data warehousing, and machine learning methods.

The authors highlight the potential of OI to address the inefficiencies of traditional silicon-based AI systems by offering faster decision-making, continuous learning, and greater energy and data efficiency. By leveraging the natural intelligence of brain organoids, OI aims to overcome the limitations of current AI technologies, which are constrained by high energy consumption and the von Neumann bottleneck. Brain organoids, with their ability to self-organize and form complex 3D neural networks, provide a more accurate model of brain function compared to traditional 2D cultures. This research also emphasizes the ethical implications of OI, advocating for an embedded ethics approach to address the potential moral and societal concerns associated with developing and utilizing brain organoids. The anticipated applications of OI extend beyond biocomputing, potentially offering new insights into the pathophysiology of developmental and degenerative diseases and informing novel therapeutic approaches. As the field progresses, the collaborative and iterative nature of the OI program aims to refine neurocomputational theories and develop scalable, high-throughput biocomputing systems, ultimately contributing to a new generation of biological computing technologies.

\vspace{-0.2cm}
\subsection{Ballav et al., 2024}
Ballav et al. \citep{Ballav_Ranjan_Sur_BasuSil_2024} explore the concept of Organoid Intelligence by integrating brain organoid technology with computational methods, aiming to bridge AI and biological computing. Brain organoids are three-dimensional lab-grown structures replicating certain aspects of the human brain's organization and function, providing a platform to study neurodevelopmental processes, disease modeling, and drug testing. The chapter delves into the potential of OI to enhance our understanding of human cognitive functions and achieve significant biological computational proficiencies. Advanced bioengineering techniques, such as high-throughput single-cell RNA sequencing, proteomics, and imaging, generate comprehensive datasets from organoids, revealing complex patterns and relationships that are analyzed through machine learning and AI. This convergence aims to create a new frontier in biocomputing and \textbf{intelligence-in-a-dish}, offering unique insights into brain function and potential therapeutic strategies for neurodevelopmental and neurodegenerative disorders.

Integrating AI with organoid technology leverages the extraordinary biological processing power of brain organoids and the computational prowess of AI to create advanced biocomputing systems. These OI systems can process complex inputs, study the physiology of learning, and generate responses to control peripheral devices. Hence, the chapter highlights the advantages of brain organoids over traditional 2D cultures and animal models, emphasizing their ability to accurately model human brain development and disease progression. Furthermore, the authors discuss the ethical implications of OI, particularly the potential for organoids to develop higher-order functions and the moral considerations surrounding their use. Ballav et al. also present potential applications of OI, spanning from advancements in understanding brain development and diseases to creating novel neuromimetic AI algorithms and brain-computer interface technologies. Despite the challenges, integrating AI and OI represents a promising new frontier in neuroscience and biocomputing, offering significant potential for scientific discovery and technological innovation.

\section{Approach}
\label{sec:approach}
\begin{figure}[H]
    \centering
    \includegraphics[width=0.9\linewidth]{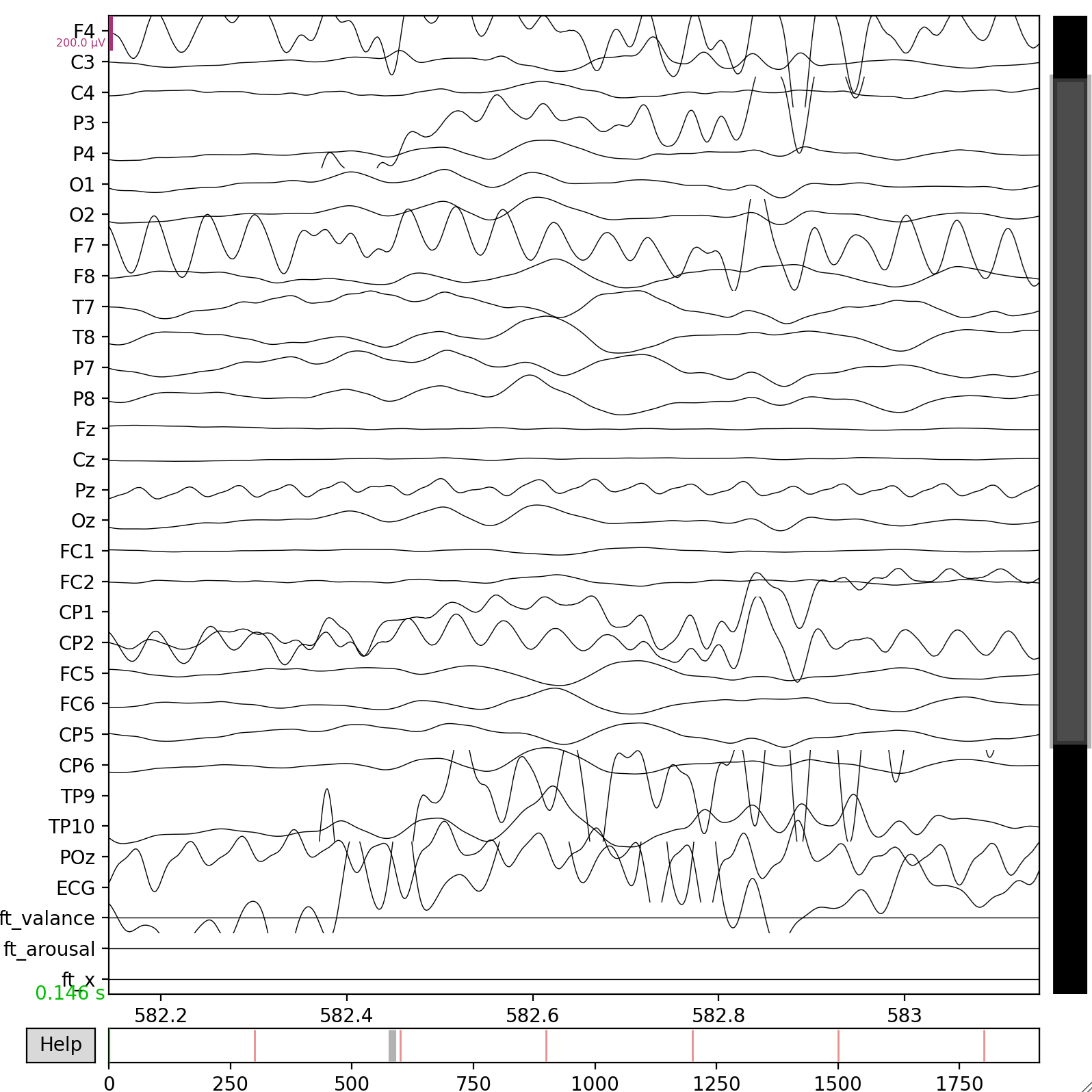}
    \caption{A sample timestep from the EEG recording of subject \#1 listening to classical piano music.}
    \label{fig:sample_eeg}
\end{figure}

Here, we detail the methodological framework utilized to simulate neural responses to classical music using an \textbf{AI/OI} model. The study employs the ``Joint EEG-fMRI recording during affective music listening'' dataset by Daly et al. \citep{Daly_Nicolaou_Williams_Hwang_Kirke_Miranda_SlawomirNasuto_2020}, focusing solely on the EEG data collected from 21 subjects exposed to seven distinct pieces of classical piano music (see \underline{\hyperref[fig:sample_eeg]{Figure \ref{fig:sample_eeg}}}). The preprocessing of EEG data involved the application of the \textbf{MNE-Python} library, which facilitated the band-pass filtering and artifact correction using Independent Component Analysis. The subsequent epoching process, driven by event timestamps from the dataset, allowed for the extraction of relevant neural responses corresponding to specific auditory stimuli. 

Following the data preprocessing, we train a \textbf{TensorFlow} model to predict EEG responses based on the audio features of the classical music pieces. The model was designed to capture the intricate relationship between auditory inputs and corresponding neural outputs, thereby creating a realistic simulation of EEG recordings. The predicted EEG data serve as a proxy for the neural activity that could be recorded from a biological brain organoid under similar conditions. This simulated data is then integrated with the \textbf{PyOrganoid} library, a novel tool developed for this research, enabling the simulation of organoid learning and response. 

\vspace{-0.4cm}
\subsection{PyOrganoid}
PyOrganoid is a novel \textbf{Python} package developed to facilitate the simulation of organoids, with a focus on studying Organoid Intelligence and Organoid Learning \citep{Szelogowski_pyorganoid_A_Python_2022}. This new library emerged from the need to integrate biological realism into computational models, thereby enhancing our understanding of neural processes and their applications in artificial intelligence and bioengineering. Our motivation for development is to provide researchers, educators, and academic institutions with a versatile and user-friendly tool that bridges the gap between biological systems and machine learning models. By simulating organoids, researchers can explore complex neural dynamics, study the effects of various stimuli on organoid behavior, and develop new computational models that mimic biological intelligence without the cost of developing the organoid ahead of time.

\begin{figure}[H]
    \centering
    \includegraphics[width=0.9\linewidth]{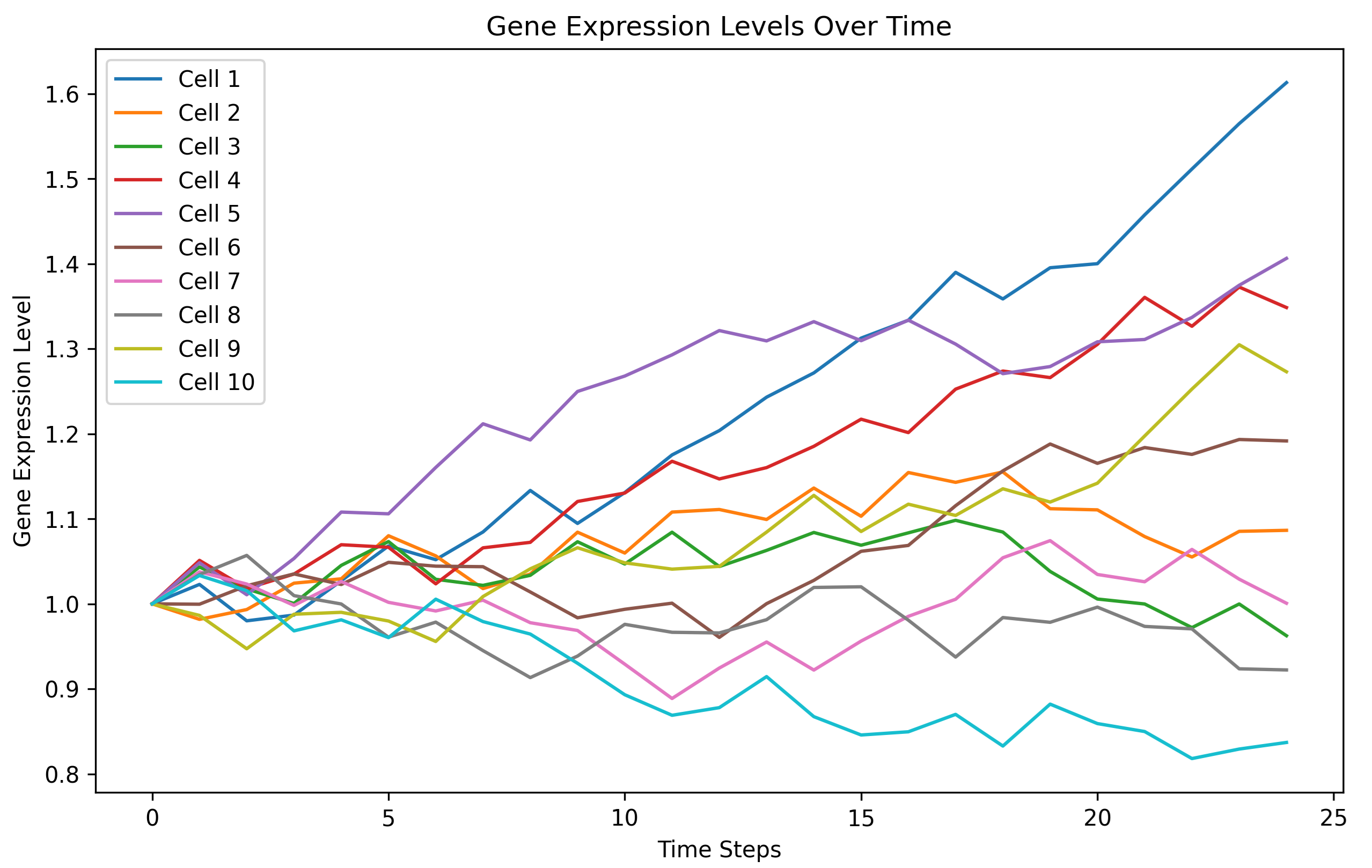}
    \caption{Simulation history from an example Gene Regulation organoid \citep{Szelogowski_pyorganoid_A_Python_2022}.}
    \label{fig:gene_exp_organoid}
\end{figure}

The library boasts a range of features designed to support diverse research applications, offering a simple and intuitive API and ensuring accessibility for users at different levels of expertise (similar to the BioDynaMo C++ library). It supports major machine learning libraries such as TensorFlow, PyTorch, Scikit-Learn, and ONNX, facilitating the integration of advanced machine learning models into organoid simulations (with a simple API for creating ad hoc adapters for other ML model types). The package includes a growing library of organoid models, each tailored to simulate different biological phenomena. Additionally, PyOrganoid provides robust visualization tools for monitoring and analyzing organoid simulations, enabling researchers to gain insights into the underlying mechanisms driving organoid behavior (see \underline{\hyperref[fig:gene_exp_organoid]{Figure \ref{fig:gene_exp_organoid}}}). Likewise, it also includes various simulation environments and scheduling algorithms, allowing users to create realistic and dynamic experimental setups.

PyOrganoid is designed to be extensible, catering to various research needs. Hence, it includes several types of OI models (as of July, 2024), each representing different aspects of biological intelligence:
\begin{itemize}
    \item Spiking Neuron Organoids
    \item Growth/Shrinkage Organoids (i.e., volumetric models)
    \item Differentiation Organoids (i.e., artificial neurons)
    \item Chemotaxis Organoids
    \item Immune Response Organoids
    \item Synaptic Plasticity Organoids (for STDP)
    \item Metabolic Organoids
    \item Gene Regulation Organoids (i.e., for gene expression)
\end{itemize}
The \textbf{environments} provided within the library \textemdash{} including Gradient, Temperature, Stochastic, Chemical Gradient, and Electric Field Environments \textemdash{} enable the simulation of diverse experimental conditions. Moreover, the library supports multiple \textbf{simulation schedulers}, such as Stochastic, Priority, and Parallel Schedulers, which help manage the execution of complex simulations. 

\section{Implementation}
\label{sec:implementation}
To normalize the EEG data from the dataset in preparation for our deep learning model, each recording underwent a preprocessing pipeline using the \textbf{MNE-Python} library. This process included the application of a band-pass filter with a frequency range of 1 to 40 Hz using the `firwin' design to remove any high-frequency noise and low-frequency drift \citep{Newman_2020}. Subsequently, \textbf{Independent Component Analysis (ICA)} was applied to correct for artifacts such as eye movements and muscle activity, ensuring cleaner and more accurate EEG signals \citep{Mayeli_Zotev_Refai_Bodurka_2016}. The preprocessed data was then segmented into epochs based on event timestamps, which were derived from the dataset and converted into sample indices corresponding to the onset of each musical piece.

Due to the expansive (file) size of the EEG dataset, a data management strategy leveraging \textbf{NumPy's} memory-mapped arrays (\textbf{memmap}) was implemented. This approach allowed the creation of an external data generator, facilitating the efficient handling and processing of large-scale data without overloading system memory \citep{NumPy}. 
Additionally, audio features were extracted from the classical music pieces using the \textbf{Librosa} library, focusing on the \textbf{Mel-Frequency Cepstral Coefficients (MFCCs)}, which are pivotal for capturing the timbral aspects of the audio signals \citep{Szelogowski_2021}. These MFCC features were then aligned with the corresponding EEG epochs to create a comprehensive dataset, integrating both neural and auditory data. This dataset thus served as the foundation for training various machine learning models in an attempt to simulate the neural responses to classical piano music.

A custom data generator was developed to facilitate the seamless integration of the extensive and complex dataset with the machine learning models. This generator utilized the extracted audio features and the preprocessed EEG data to create batches of input-output pairs (respectively) for model training. By shuffling indices and ensuring proper alignment between the EEG epochs and audio feature segments, the generator maintained data consistency and enabled efficient batch processing. As well, the use of NumPy's memmap and a well-structured data generator not only optimized memory usage but also ensured scalability and robustness in handling large datasets. 

\vspace{-0.2cm}
\subsection{AI Model}
\begin{figure}[h]
    \centering
    \includegraphics[width=0.785\linewidth]{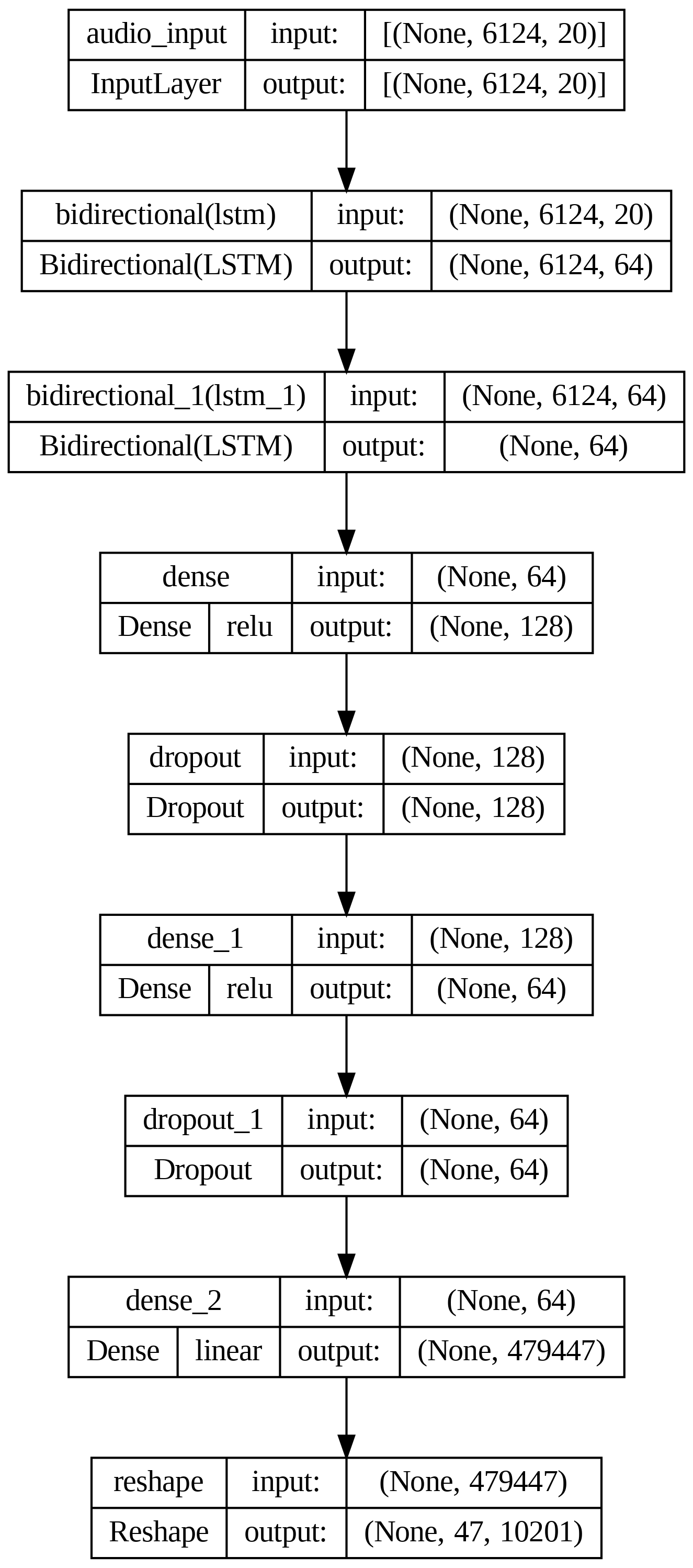}
    \caption{Bidirectional LSTM Network, created with Tensorflow for OI interfacing.}
    \label{fig:bilstm}
\end{figure}

The development and design of the AI model for this study centered on a \textbf{Bidirectional Long Short-Term Memory (Bi-LSTM)} network, which emerged as the most effective architecture compared to LSTM, Convolutional Neural Network (CNN), Recurrent Neural Network (RNN), and Gated Recurrent Unit (GRU) models. This architecture was chosen for its ability to capture temporal dependencies in both forward and backward directions, enhancing the model's capacity to understand the complex relationships within the audio features and corresponding EEG data \citep{Szelogowski_Mukherjee_Whitcomb_2022}. The input to the model consisted of the MFCCs extracted from the audio recordings, which are known for their efficacy in representing the short-term power spectrum of sound. Likewise, the model's output was designed to predict the EEG signals, comprising 47 values \textemdash{} each corresponding to a specific EEG node from the dataset (see \underline{\hyperref[fig:bilstm]{Figure \ref{fig:bilstm}}}). As such, this setup allowed the model to learn a comprehensive mapping from audio features to neural responses, leveraging the temporal information encapsulated in the Bi-LSTM layers.

To ensure the model's optimal performance, a grid search was conducted to fine-tune hyperparameters, including the number of LSTM units, dropout rates, recurrent dropout rates, and regularization parameters. The architecture comprised an initial input layer accepting the MFCCs, followed by two Bidirectional LSTM layers with 64 units each, incorporating dropout and recurrent dropout to mitigate overfitting and enhance generalization. These layers were succeeded by dense layers with ReLU activation and L2 regularization, further refining the learned features before the final output layer. The output layer was configured to match the dimensionality of the EEG signals, with a linear activation function to facilitate accurate regression of the neural responses. Training utilized the \textbf{Mean Squared Error (MSE)} loss function, with \textbf{Mean Absolute Error (MAE)} as an additional metric to monitor model performance. The optimization was carried out using the Adam optimizer with an exponential decay learning rate schedule, complemented by callbacks such as early stopping, learning rate reduction, and model checkpointing to ensure robust training and prevent overfitting. 

\subsection{Pianoid - OI Model}
\begin{figure}[H]
  \begin{adjustbox}{addcode={\begin{minipage}{\width}}{\caption{%
      EEG-Response Organoid with 47 cells, connected to the Tensorflow Bi-LSTM model and an Audio-based Environment.
      \label{fig:eegorganoid}  
      }\end{minipage}},rotate=90,center}
      \includegraphics[scale=.085]{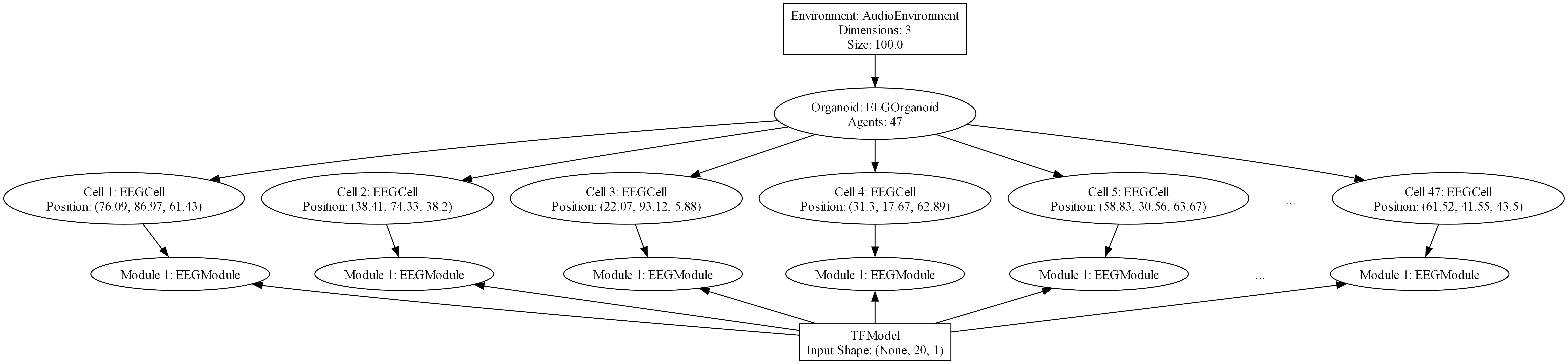}%
  \end{adjustbox}
\end{figure}

Our ``Organoid Learning'' model \textemdash{} \textbf{Pianoid} \textemdash{} was designed and developed to simulate neural responses to classical piano music, utilizing the output from the AI model that predicts EEG signals. The model contains 47 cells, each representing one of the 47 EEG channels from the dataset (see \underline{\hyperref[fig:eegorganoid]{Figure \ref{fig:eegorganoid}}}). These cells were instantiated within the \textbf{PyOrganoid} framework, with each cell initialized at random positions within a three-dimensional environment (akin to a 3D brain organoid). Each cell was associated with a corresponding channel index and equipped with a module designed to process the input data provided by the AI model. This setup allowed for a detailed and localized simulation of neural activity, where each cell's behavior mimicked the dynamics of an individual EEG signal.

The audio-based environment facilitates the interaction between the organoid and the auditory stimuli. It was designed to process audio files, extracting relevant features such as MFCCs, which were then used as inputs to the organoid model. The \textbf{AudioScheduler}, an ad hoc class designed specifically for this research, orchestrated the simulation by ``playing'' the audio to the organoid over discrete time steps. This scheduler fed the extracted audio features into the trained AI model, which then predicted the corresponding EEG signals. These predicted signals were distributed to the respective cells, updating their states (once per timestep) to reflect the neural responses to the music being played.

Several custom classes were developed to demonstrate the simplicity and extensibility of the PyOrganoid library. These included the \textbf{AudioEnvironment}, which handled the loading and processing of audio files; the \textbf{EEGCell}, representing individual EEG channels; the \textbf{EEGModule}, responsible for interfacing with the machine learning model; and the \quad\quad\quad \textbf{EEGOrganoid}, which integrated these components into a cohesive unit. This modular and extensible approach allowed for straightforward experimentation and adaptation, enabling other researchers to easily extend the PyOrganoid framework for various use cases. The scheduler simulated the auditory environment in real-time, updating each cell's activity based on the AI model's output, creating a dynamic and interactive simulation that closely mirrored the neural processing of auditory stimuli.

\clearpage

\section{Evaluation}
\label{sec:evaluation}
\begin{figure}[h]
    \centering
    \includegraphics[width=1\linewidth]{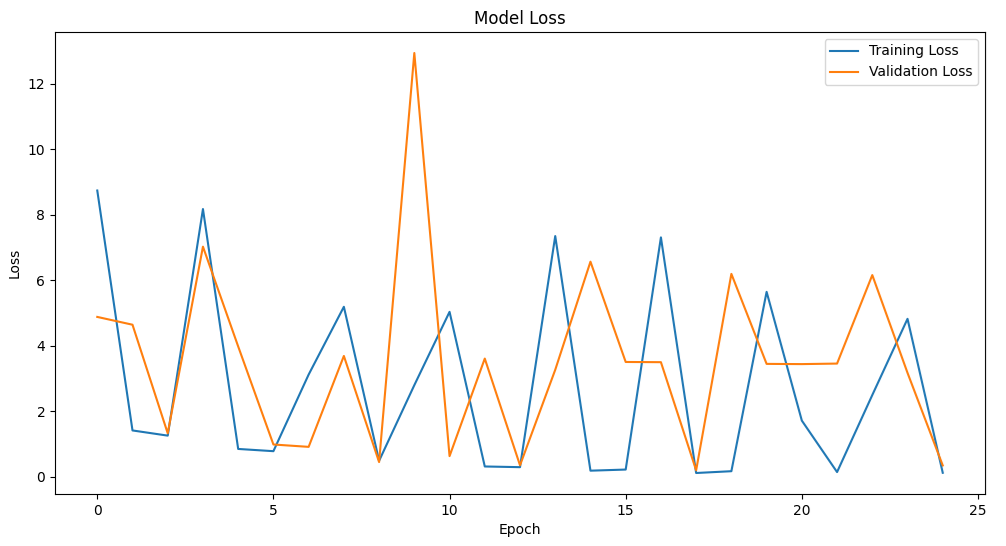}
    \caption{Bi-LSTM training and validation loss history.}
    \label{fig:lstm_loss}
\end{figure}

\begin{figure}[h]
    \centering
    \includegraphics[width=1\linewidth]{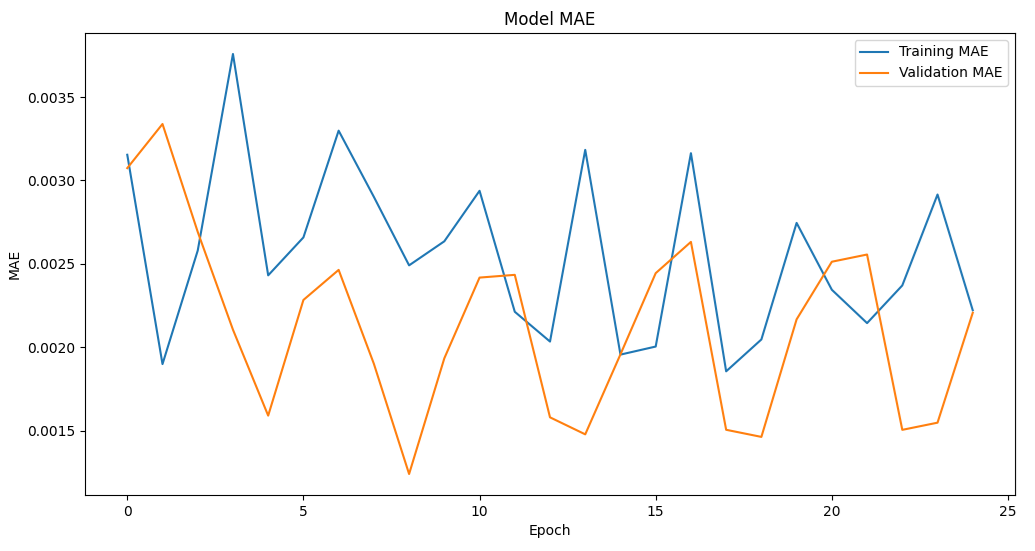}
    \caption{Bi-LSTM training and validation MAE history.}
    \label{fig:lstm_mae}
\end{figure}

The evaluation of the AI model's performance involved several metrics and comparative analyses to ensure the accuracy and reliability of the simulated EEG signals. One primary metric used was the \textbf{Mean Absolute Error (MAE)}, which measures the average magnitude of errors between the predicted and actual EEG values. The training and validation \textbf{Mean Squared Error (MSE)} loss curves (see \underline{\hyperref[fig:lstm_loss]{Figure \ref{fig:lstm_loss}}}) reveal significant fluctuations throughout the training process, indicating potential variability in the model's learning stability \citep{Antelis_Montesano_Ramos-Murguialday_Birbaumer_Minguez_2013}. However, the overall trend suggests a gradual convergence, with the final epochs showing a reduction in both training and validation loss. This behavior was similarly reflected in the MAE curves (see \underline{\hyperref[fig:lstm_mae]{Figure \ref{fig:lstm_mae}}}), which display the differences in mean absolute error for both training and validation datasets over the epochs. Despite occasional spikes in the validation MAE, the model generally maintained a stable MAE, suggesting its capacity to generalize the learned features from the training data to unseen validation data. Likewise, the most accurate checkpoint of the model (our current weights) attained a validation loss of 0.3443 and MAE of 0.0015; hence, we believe these to be sufficient for the purpose of this simulation.

\vspace{-0.15cm}

\vspace{-0.15cm}
\clearpage

\begin{figure}[H]
    \centering
    \includegraphics[width=0.8\linewidth]{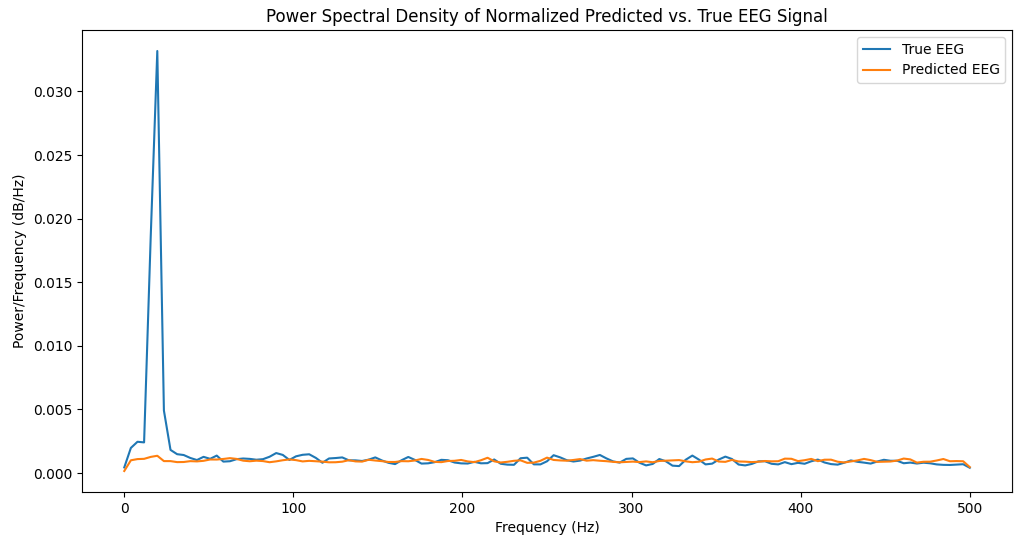}
    
    \vspace{-0.25cm}
    \caption{Comparison of (normalized) predicted versus actual EEG signal PSD.}
    \label{fig:lstm_psd}
\end{figure}

\vspace{-0.65cm}
\begin{figure}[H]
    \centering
    \includegraphics[width=0.8\linewidth]{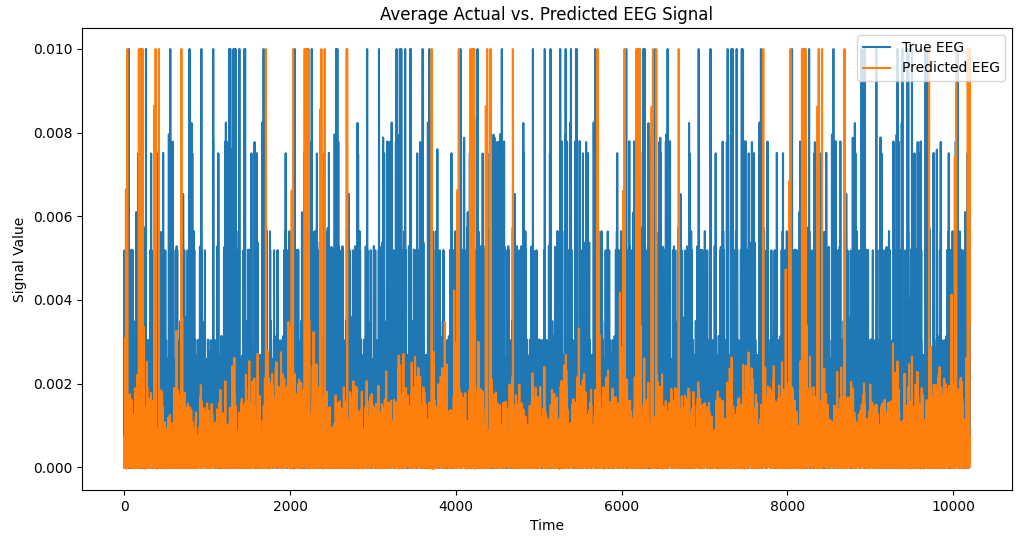}
    
    \vspace{-0.25cm}
    \caption{Comparison of average predicted versus \quad\quad actual EEG signal (across all 47 channels).}
    \label{fig:lstm_avgeeg}
\end{figure}

\vspace{-0.3cm}
A critical aspect of the evaluation was assessing the closeness of the model's output to actual EEG recordings, particularly regarding spectral characteristics. The \textbf{Power Spectral Density (PSD)} comparison between the predicted and true EEG signals (see \underline{\hyperref[fig:lstm_psd]{Figure \ref{fig:lstm_psd}}}) provided insights into how well the model captured the frequency components of the EEG signals \citep{Wang_Wang_Yu_Wei_Yang_Deng_2015}. Plotting the PSD revealed that while the model could replicate the general shape of the EEG spectrum, including the dominant peaks, there were discrepancies in the amplitude of these peaks. This suggests that while the model can learn the temporal dynamics to some extent, there is still a gap in accurately representing the full spectral characteristics of the EEG signals. Additionally, the average EEG signal comparison (see \underline{\hyperref[fig:lstm_avgeeg]{Figure \ref{fig:lstm_avgeeg}}}) further highlighted these differences, with noticeable deviations between the predicted and actual signal amplitudes and phases over time.

These evaluation results highlight the complexity of simulating EEG responses to musical stimuli and highlight areas for future model refinement. The observed fluctuations in loss and MAE, coupled with the discrepancies in spectral representations, suggest that the model might benefit from additional regularization techniques, more sophisticated architecture adjustments, or enhanced feature extraction methodologies. Moreover, the use of correlation coefficients between the predicted and actual EEG signals served as a quantitative measure of the model's predictive accuracy, with the current model achieving a moderate correlation ($r \approx 0.51$). This indicates partial success in capturing the essential features of the EEG responses but also points to the need for further improvement. Overall, the evaluation emphasizes the challenges in modeling complex neural responses and the potential pathways for enhancing the fidelity of such simulations in the context of organoid intelligence research.

\section{Discussion}
\label{sec:discussion}
\begin{figure}[H]
    \centering
    \includegraphics[width=0.875\linewidth]{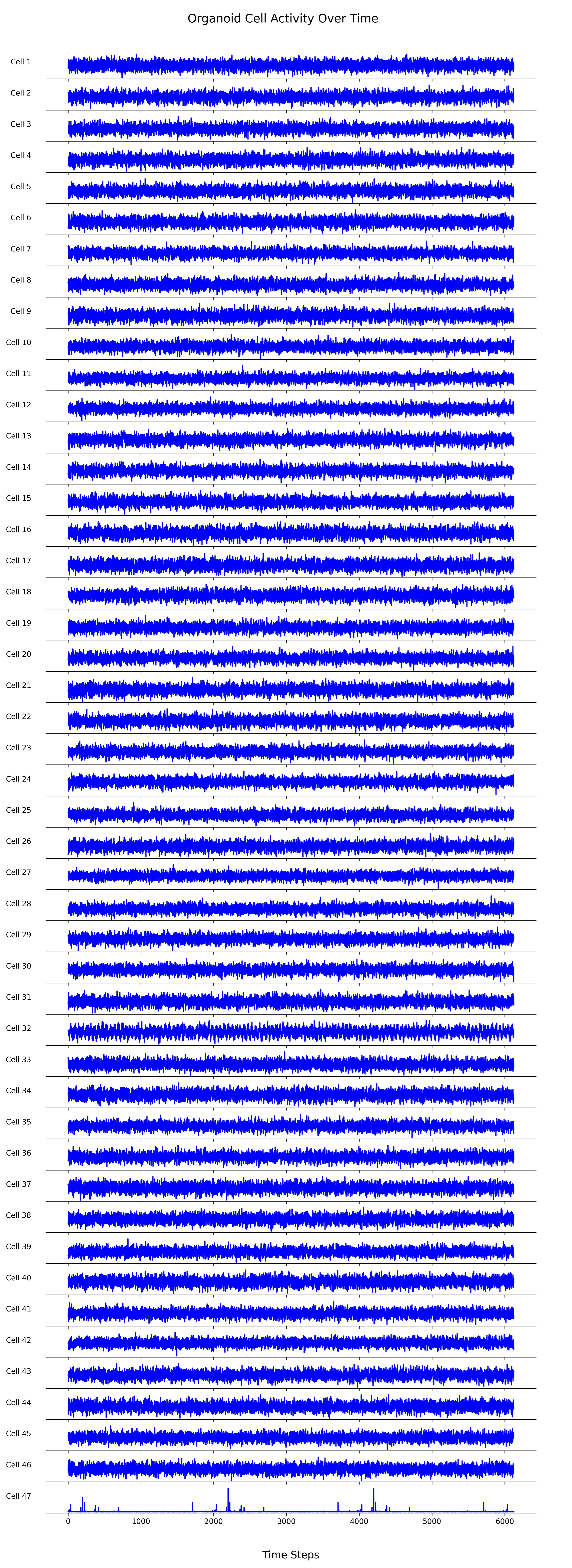}
    \caption{EEG signal analysis from Pianoid model across all time steps (listening to the first piece in the dataset, Chopin's \textit{Revolutionary Étude}, Op. 10 No. 12).}
    \label{fig:organoid_eeg}
\end{figure}

The successful simulation of EEG signals in response to classical music using the \textbf{Pianoid} model demonstrates a significant advancement in organoid intelligence research. By leveraging a Bidirectional LSTM network, the model effectively predicted EEG signals from the audio features of classical music, showcasing the potential of using computational models to emulate complex neural responses. The EEG signal plot (see \underline{\hyperref[fig:organoid_eeg]{Figure \ref{fig:organoid_eeg}}}) illustrates the activity across the 47 simulated EEG channels, revealing distinct patterns that correlate with the musical input, thereby validating the model's capability to generate plausible neural responses. This work thus highlights the feasibility of integrating organoid models with advanced machine learning techniques to study human cognitive processes, such as music perception and cognition. 

The development and use of the \textbf{PyOrganoid} library played a crucial role in achieving these results, offering a versatile and extensible framework for simulating organoid behavior and interfacing with machine learning models. Its modular design and support for multiple machine learning frameworks enabled the seamless integration of the Bi-LSTM model with the organoid simulation. This facilitated a controlled environment for testing hypotheses about neural responses to music and provided a platform for future research in organoid intelligence. However, the library also has limitations, including its dependency on predefined models and environments, which may not fully capture the complexity and variability of biological systems. Likewise, while PyOrganoid can simulate aspects of neural activity, it lacks the biochemical and structural nuances of actual brain organoids, limiting its ability to replicate the full spectrum of neural responses observed in living tissue.

Despite the promising outcomes, the study also highlights significant challenges and limitations. Though adequate for demonstrating feasibility, the model's predictive accuracy revealed discrepancies in the power spectral density and correlation measures between predicted and true EEG signals. These issues emphasize the need for more sophisticated models capable of capturing the full complexity of neural activity, potentially through more advanced architectures like transformer models or real-time feedback mechanisms. Additionally, the lack of actual biological components in PyOrganoid simulations poses a limitation in fully understanding the dynamic and adaptive nature of real neural networks. Future research should aim to address these limitations by integrating more biologically inspired elements and expanding the range of stimuli and individual variability in the dataset. This approach could lead to more accurate and comprehensive simulations, offering deeper insights into the neural mechanisms underlying human cognition and opening new avenues in neuroengineering and personalized medicine.

\vspace{-0.25cm}
\section{Conclusion}
\label{sec:conclusion}
Our research successfully demonstrates the potential of using OI/OL models to simulate neural responses to classical music (in this case, the Pianoid model). By employing a Bidirectional LSTM network integrated within the PyOrganoid framework, the study successfully predicted EEG signals corresponding to the auditory features of classical music, thereby providing a novel approach to studying music perception and cognition. The accuracy of the model's predictions, as reflected in the generated EEG signal plots, indicates a promising step towards developing computational tools that can further emulate complex neural processes. 

The primary contributions of this research include the creation of the PyOrganoid library, which offers a robust and extensible framework for simulating organoid behavior and interfacing with various machine learning models. Moreover, the success of the Pianoid model in generating EEG-like signals provides valuable insights into the neural mechanisms underlying music cognition, paving the way for further exploration in this domain \citep{Curzel_Tillmann_Ferreri_2024}. These findings reinforce the potential of organoid intelligence as a complementary tool to traditional neurobiological methods, enabling more controlled and reproducible studies of brain function. The implications of this work also extend beyond the immediate study of music perception, offering insights into the broader field of synthetic biology and its applications in understanding and replicating human cognitive functions. 

\vspace{-0.25cm}
\subsection{Future Work}
Future researchers could explore several promising directions to enhance the accuracy and applicability of the Pianoid model and the PyOrganoid framework, including augmenting/extending the audio dataset (which could help refine the model's ability to predict EEG responses more accurately across diverse auditory stimuli \citep{Mignot_Peeters_2019}). Additionally, implementing more sophisticated or hybrid deep learning architectures, such as combining CNNs with recurrent layers, may improve the model's capacity to capture complex temporal and spectral features of the music \citep{Golmohammadi_Harati_Nejad_Torbati_Lopez_de_Diego_Obeid_Picone_2019}. Developing a more robust and biologically plausible organoid model, potentially incorporating advanced cellular behaviors and interactions, could also enhance the fidelity of simulated neural responses.
These advancements would not only improve the scientific utility of PyOrganoid but also expand its potential applications in cognitive neuroscience, \textbf{bioinformatics}, and the development of cutting-edge neurotechnological interfaces.

\bibliographystyle{ACM-Reference-Format}
\bibliography{references}

\end{document}